\documentclass{article}

\usepackage[preprint]{neurips_2024}

\usepackage[utf8]{inputenc} 
\usepackage[T1]{fontenc}    
\usepackage{hyperref}       
\usepackage{url}            
\usepackage{booktabs}       
\usepackage{amsfonts}       
\usepackage{nicefrac}       
\usepackage{microtype}      
\usepackage{xcolor}         
\usepackage{times}
\usepackage{soul}
\usepackage{url}
\usepackage{graphicx}
\usepackage{amsmath}
\usepackage{amsthm}
\usepackage{booktabs}
\usepackage{algorithm}
\usepackage{algorithmic}
\usepackage[switch]{lineno}
\usepackage{wrapfig}

\usepackage{amsmath,amsfonts}
\usepackage{array}
\usepackage[caption=false,font=normalsize,labelfont=sf,textfont=sf]{subfig}
\usepackage{textcomp}
\usepackage{stfloats}
\usepackage{url}
\usepackage{verbatim}
\usepackage{multirow}
\usepackage{tabularx}
\usepackage{amssymb}
\usepackage{subcaption}

\title{Neural Assembler: Learning to Generate Fine-Grained Robotic Assembly Instructions from Multi-View Images}

\author{%
  Hongyu Yan \\
  Peking University\\
  \texttt{pku\_wdyhy@pku.edu.cn} \\
  \And
  Yadong Mu \\
  Peking University\\
  \texttt{myd@pku.edu.cn} \\
}

\begin{document}

\maketitle

\begin{abstract}
  Image-guided object assembly represents a burgeoning research topic in computer vision. This paper introduces a novel task: translating multi-view images of a structural 3D model (for example, one constructed with building blocks drawn from a 3D-object library) into a detailed sequence of assembly instructions executable by a robotic arm. Fed with multi-view images of the target 3D model for replication, the model designed for this task must address several sub-tasks, including recognizing individual components used in constructing the 3D model, estimating the geometric pose of each component, and deducing a feasible assembly order adhering to physical rules. Establishing accurate 2D-3D correspondence between multi-view images and 3D objects is technically challenging. To tackle this, we propose an end-to-end model known as the Neural Assembler. This model learns an object graph where each vertex represents recognized components from the images, and the edges specify the topology of the 3D model, enabling the derivation of an assembly plan. We establish benchmarks for this task and conduct comprehensive empirical evaluations of Neural Assembler and alternative solutions. Our experiments clearly demonstrate the superiority of Neural Assembler.
\end{abstract}

\section{Introduction}

The assembly task necessitates predicting a sequence of operations for the placement of various components. Accurate and efficient assembly algorithms play a pivotal role in robotics. These assembly challenges are pervasive in daily life, as in scenarios like constructing LEGO models \cite{chung2021brick}, assembling furniture \cite{suarez2018can}, and Minecraft \cite{chen2019order}. In previous research, Chen et al. \cite{chen2019order} suggested replicating human building order with spatial awareness to construct Minecraft houses without target information. Wang et al. \cite{wang2022translating} introduced a step-by-step approach for assembling LEGO models based on the assembly manual, while the work in \cite{zhan2020generative} focused on predicting the 6-DoF pose of each component based on the object's class for assembly. Li et al. \cite{li2020learning} predicted the poses of parts using a single image.

\begin{figure*}[t]
\begin{center}
\includegraphics[width=\linewidth]{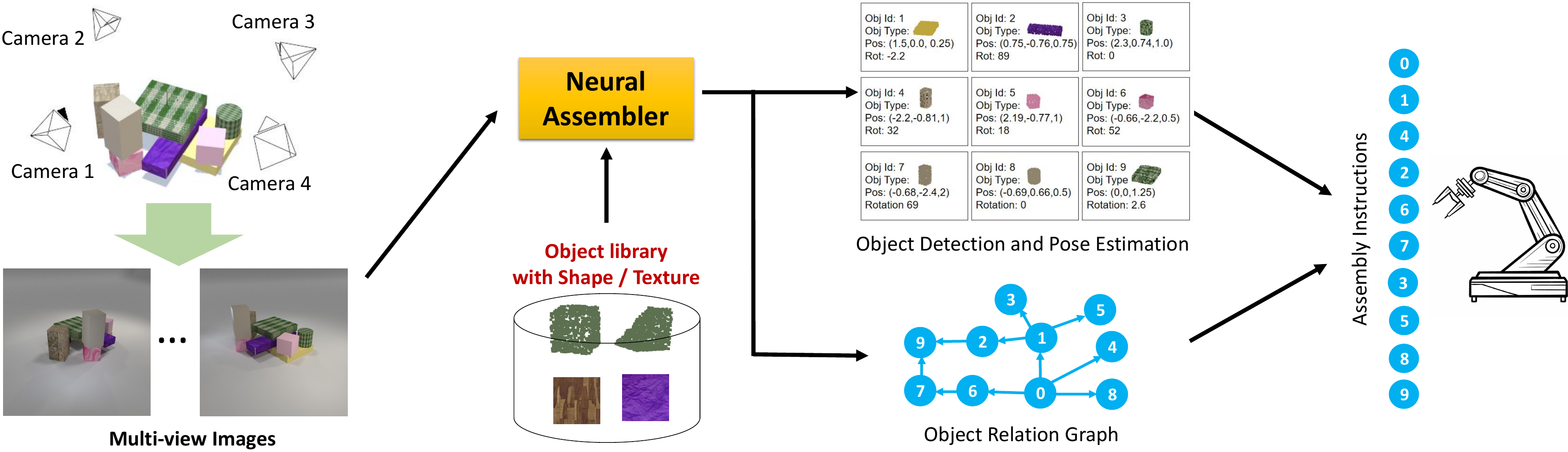}
\end{center}
   \caption{\small Schematic illustration of the proposed Neural Assembler. See Section 3 for more details.}
\label{fig:idea}
\end{figure*}

In this study, we define a new task of image-guided assembly. We are provided with a set of multi-view images captured from a 3-D model, assuming it is built with components from a pre-specified library. The goal of the task is to generate a sequence of fine-grained assembly instructions, encompassing all parameters—such as component types, geometric poses of each component, and assembly order—in accordance with physical rules and suitable for execution by a robotic arm. The task serves as a valuable testbed for advancing vision-guided autonomous systems, presenting a range of technical challenges. Firstly, understanding the correspondence between 2D images and 3D objects is crucial. Given that certain components in the 3D model might be entirely obscured from specific viewpoints, we employ multi-view images (e.g., typically 4 in this study) as input. The algorithm must effectively integrate information from images captured from multiple perspectives. Secondly, estimating critical information for each component is non-trivial. With a 3D library in place, the algorithm needs to segment components across all images and categorize each based on predefined object types in the library, mainly using shape and texture cues. In addition, one also needs estimate the 3-D spatial position and rotation matrix of each component. Thirdly, obtaining typological information for all components is necessary to formulate a physically feasible assembly plan.

Importantly, observations of a component in images are often incomplete, primarily due to frequent occlusions. This poses a substantial challenge in fully understanding and interpreting the scene. Such occlusions are particularly challenging when assembling complex, multi-layerd models. For this novel task, we propose an end-to-end neural network, dubbed as Neural Assembler. The computational pipeline of Neural Assembler is illustrated in Figure~\ref{fig:idea}. Taking multi-view images and a 3-D component library as input, Neural Assembler not only identifies each component from images but also determines its 3D pose at each step of assembly. Leveraging images from multiple viewpoints, our method aims to enhance the overall scene understanding and accurately predict the order for placing parts in assembly tasks.

We present two datasets for the proposed image-guided assembly task, namely the CLEVR-Assembly dataset and LEGO-Assembly dataset. Each sample in these datasets comprises images of the object captured from various perspectives, along with the pose of each part (2D keypoints, mask, 3D position, and rotation), and the relationship graph of all components. Comprehensive experiments are conducted on both datasets. Due to the absence of prior work addressing this novel setting like Neural Assembler, we establish two robust baselines for comparison. The evaluations unequivocally demonstrate that Neural Assembler outperforms the baselines across all performance metrics.

\section{Related work}
\textbf{Assembly algorithms}
Assembly tasks have utilized computational strategies for part selection and placement, including action scoring \cite{yuille2006vision,bever2010analysis}, genetic algorithms \cite{lee2015finding}, and voxel-based optimization \cite{KimJ2020neuripsw,van2015part}. Manual-driven approaches have been investigated \cite{10.1145/2980179.2982416,wang2022translating,chen2019order,walsman2022break,chung2021brick}, with LSTM showing limitations in long sequence predictions \cite{walsman2022break}, and the reinforcement learning method\cite{chung2021brick} struggling with block complexity. Existing research into part-based 3D modeling \cite{zhan2020generative,mo2019structurenet,li2020learning,niu2018im2struct} and parts retrieval from 3D meshes \cite{chaudhuri2010data,shen2012structure,sung2017complementme} assumes prior part knowledge. Our novel task of reconstructing 3D models from multi-view images without prior information presents a unique challenge not yet addressed by none of above works.

\textbf{Multi-view scene understanding}
Many scene understanding tasks are intractable in single-view and can only be solved given multi-view inputs.
For example, The SLAM (simultaneous localization and mapping) task requires reconstructing the 3D geometric scene and estimating camera poses given a sequence of video frames. 3D-SIS~\cite{hou20193d} performed instance segmentation by reading a set of RGB-D images. MVPointNet~\cite{jaritz2019multi} was based on a set of RGB-D images and point clouds for instance segmentation. \cite{murez2020atlas}~performed 3D semantic segmentation and reconstruction based on multiple RGB images. Similarly in our work, the algorithm understands the structure of 3D brick models and predicts the poses of parts based on multi-view images.

\textbf{Structural modeling for scenes or 3-D models}
Recent studies have focused on inferring structural representations like graphs \cite{johnson2015image,cong2023reltr,li2022sgtr} and programmatic descriptions \cite{ellis2018learning,liu2019learning,wu2017learning} from images. Techniques range from encoder-decoder architectures for triplet generation \cite{cong2023reltr} to efficient graph-based methods \cite{li2022sgtr}, CNNs for primitive shape programs \cite{ellis2018learning}. An encoder-decoder approach for scene re-rendering \cite{wu2017learning} and transformers for structural change detection \cite{qiu2023graph} have also been proposed. Our work differs by using structural representations to predict assembly sequences of object parts.

\section{Neural Assembler}

\subsection{Task Specification} 

Given several multi-perspective images of a 3-D brick model $\{I_{k}\}_{k=1}^{K}$, the corresponding camera parameters with respect to some reference coordinate system, and a predefined 3-D brick (or termed as component, as used exchangeably in this paper) library $Lib = \{b_1, b_2, \ldots, b_M\}$, our algorithm identifies the bricks present in the scene, predicts each brick's pose and constructs a relational graph $G = \{V, E\}$. The vertex set $V$ correspond to the bricks and the directed edge set $E$ represent spatial configuration that can further be used to derive the assembly instructions. In particular, each brick $b_i$ in the library is denoted as $(S_i,\ T_i)$ where $S_i$ is assumed to be point clouds and $T_i$ is represented as texture images. In the relationship graph,  each node $v_i\in V$ describes the $i$-th brick information $v_i=(S_i, T_i, Kps_i, Rot_i, M_i)$ where $Kps_i\in ([0,1]\times [0,1])^{K\times 2}$ encodes the 2D keypoints (we use this notation to refer to the planar projection of the brick center) in $K$ views, $Rot_i\in  [0,\ 2\pi]^K$ are the rotation angles in $K$ views and $M_i\in \{0,1\}^{K\times H\times W}$ are the binary masks of the brick in $K$ views. The edge $e_{i,j}\in E$ explicitly describes the assembly order, where $e_{i,j}=1$ only when a brick $v_j$ is placed into the 3-D model after brick $v_i$.

\begin{figure*}[t]
  \centering
    \includegraphics[width=\linewidth]{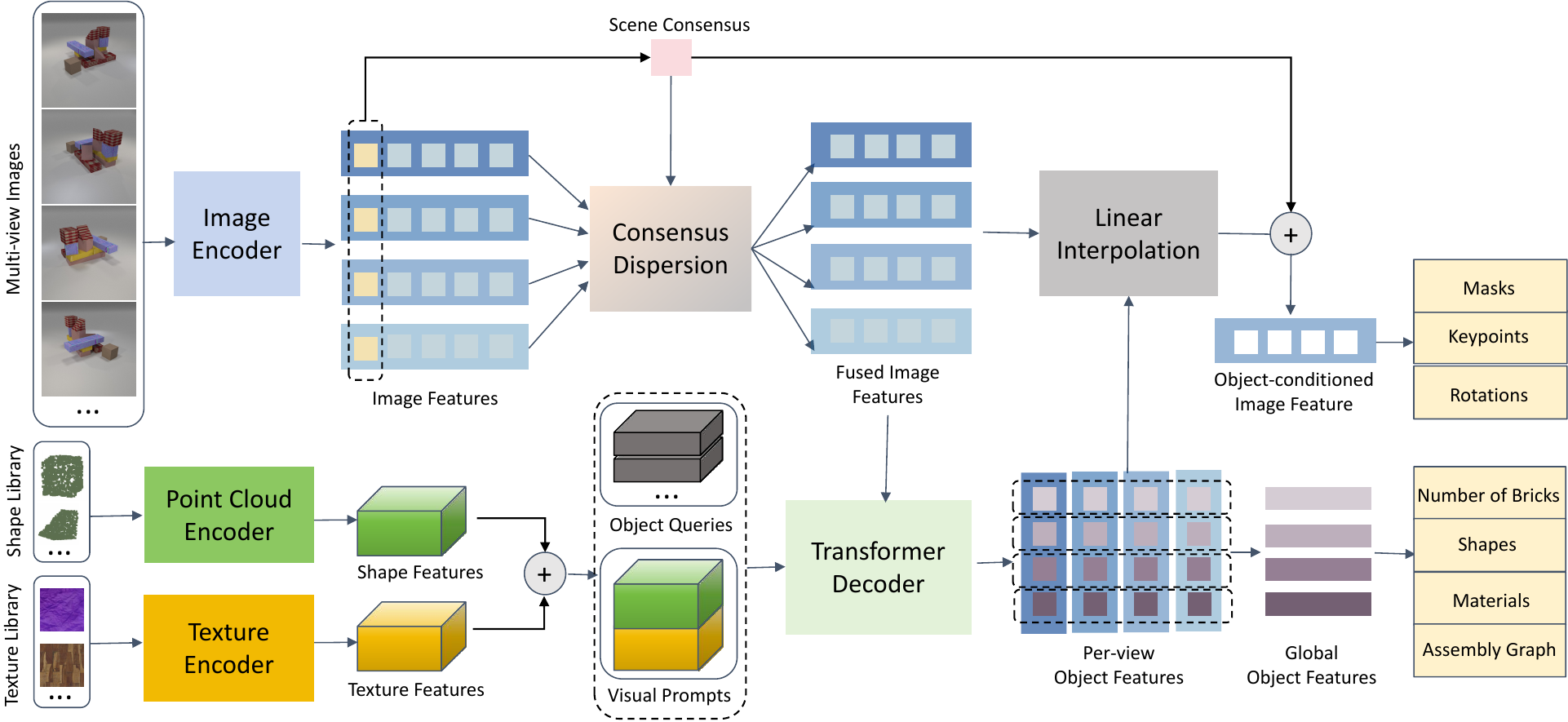}
  \hfill
  \caption{The proposed Neural Assembler architecture. An image encoder outputs the visual embeddings of multi-view images. The shape and texture library are provided as visual prompts for object detection. Then the transformer decoder module is applied to get the library-based object features. Finally, the object-conditioned image features are decoded to the bricks' masks, keypoints, and rotation angles, while the global object features are decoded to the bricks' textures, shapes, the number of blocks, and the assembly graph.}
  \label{fig:pipeline}
\end{figure*}

Here we develop a baseline for the proposed task, namely Neural Assembler, whose computational pipeline is shown in Figure~\ref{fig:pipeline}, Our model leverages a graph convolutional network (GCN) to delineate the assembly sequence effectively. The 3-D poses of objects are inferred from 2D image data across multiple views, exploiting the geometric constraints provided by the camera parameters to ensure spatial consistency in the 3D domain. This framework showcases the capability of capturing complex relational patterns and translating them into a structured assembly protocol.

\subsection{Multi-View Feature Fusion}

We adopt the pretrained CLIP image encoder to get the feature maps $F^{k}$ and CLIP feature vectors $v_{CLIP}^{k}$, $k\in\{1,2,...K\}$. Then the features of all views are fused with others, using an adapted implementation of \emph{group-based semantic agreement Transformer}~\cite{xu2023co}. The scene consensus (denoted as a vector $g$) of the images from different perspectives are extracted according to the rule: $g\ =\ \frac{1}{K}\sum_{k=1}^{K} Norm(v_{CLIP}^{k})$,
where $Norm(\cdot)$ is the $L_2$ normalization. Afterwards, the scene consensus $g$ is dispersed to multi-view image features through channel-wise multiplication:$\hat{F}^{k}=F^{k}\cdot g, \quad k=1,2, \ldots, K.$

\textbf{Template as visual prompts}. Let the texture image $T_{i}\in\mathbb{R}^{H_{T_i}\times W_{T_i} \times 3}$, where $(H_{T_i},\ W_{T_i})$ is the texture image size, represent the texture template and point clouds $S_i \in \mathbb{R}^{N_P\times 3}$ represent the shape template. In all experiments of this work, we fix the parameter $N_P$ to be all 1024, striking a balance between 
compute expense and capability of representing the bricks. For the texture image, we set both $H_{T_i}$ and $W_{T_i}$ to 224. A CNN backbone (\textit{e.g.}, ResNet-18~\cite{he2016deep}) generates template features $T_{i} \leftarrow CNN(T_i)$,
and a pointnet backbone (\textit{e.g.}, PointNet~\cite{Qi_2017_CVPR}) generates shape features $S_i \leftarrow PointNet(S_i)$.


\textbf{Transformer decoder}. The decoder accepts as input the shape feature library $\textbf{S}=\{S_i\}$, the texture feature library $\textbf{T}=\{T_i\}$ and the fused image features $\hat{F}^{k}$. As shown in Fig.~\ref{fig:pipeline}, the transformer decoder is designed to obtain the image conditioned object query.

The elements from the shape feature library $\textbf{S}$ or the texture feature library $\textbf{T}$ and the N learned object queries $O$ are concatenated to define a query $O' = [\textbf{S}, \textbf{T}, O]$.
Given the fused image features $\hat{F}^{k}$ and the query $O'$, the library-based object queries $f_i^{k}$ $(i=1,2,\ldots,N,\ k=1,2, \ldots, K)$ are obtained through a couple of decoding layers that employ the structure of FS-DETR~\cite{bulat2023fs}. Then we use an averaging operation to obtain a unified, multi-view feature representation for each object: $f_i^{global} = \sum_{k=1}^{K} f_i^{k}, \quad i=1,2,\ldots,N.$
This approach effectively integrates diverse perspectives to enhance the overall understanding for the object's attribute.

As shown in Fig.~\ref{fig:pipeline}, for the Consensus Dispersion module and Linear Interpolation module, to obtain the object-conditioned image features, we adapt the combination of CLIPSeg~\cite{luddecke2022image} and group-based semantic agreement Transformer~\cite{xu2023co}. The node features $f_i^{k}$ serve as conditional prompts. Feature fusion between node features and image features is achieved through linear interpolation. Formally, in each perspective, the object-conditioned image features $F_{i}^{k} = LI(\hat{F}^k, f_{i}^k)$
where $i=1...N,k=1,2,...K$, $LI$ is linear interpolation.
Next, the scene consensus $g$ is dispersed to multi-view object-conditioned image features through channel-wise multiplication $\hat{F}_{i}^{k} = F_i^{k}\cdot g.
$

\subsection{Brick Number Prediction}
For each view $k = 1,2,\ldots,K$, we average the N object-conditioned image features $\hat{F}_{avg}^{k} = \frac{1}{N}\sum_{i=1}^N \hat{F}_i^k$. 
Then the scene feature can be obtained by $\hat{F}_{scene} = \frac{1}{K}\sum_{k=1}^K \hat{F}_{avg}^{k}$. Finally a couple of convolution layers are used to predict the number of bricks in the scene.

\subsection{Relation Graph Generation}
Predicting the assembly sequence is equivalent to predicting the connections of bricks which can be described using the relationship graph. If brick $A$ is placed on the top of $B$, then there is a directed edge from $B$ to $A$. A graph convolutional network~\cite{kipf2016semi} is adopted to predict the existence of each edge in the graph.
Given the complete graph $G = (V, E)$ with initial node features $p_{i}^0\ =\ f_i^{global}$. Similar to~\cite{johnson2018image,wald2020learning}, we implemented the GCN using the MLP (multi-layer perception) structure. Each iteration of message passing starts from computing edge features: $
e_{i,j}^{t+1} \leftarrow MLP([p_{i}^t, p_{j}^t])$. An average-pooling is performed over all edge features connected to a node, obtaining an updated node feature
\begin{eqnarray}
p_{i}^{t+1}=\frac{1}{|\{u|(u,i)\in E\}|}\sum_{(u,i)\in E}e_{u,i}^{t+1}.
\end{eqnarray}

After gathering the edge features via graph convolutions as $\{e_{i,j}^{t}\}_{t=1}^T$, we use another MLP to predict the probability of the existence of each edge, $P_{(i,j)}=MLP(e_{i,j}^T)$

Finally, to determine the assembly sequence during inference, the directed edges are sorted in descending order according to the predicted probability and subsequently incorporated into the directed graph. If a loop is formed after adding an edge, this edge will not be added. The process continues until it reach a state where there exists a vertex that can reach all other vertices in the graph.

\subsection{Pose Estimation}


The $N$ object query features $\{f_i^{global}\}_{i=1}^N$ are used for shape and texture classification. The shape prediction head predicts the shape label and the texture prediction head predicts the texture label.

\textbf{Mask and heatmap prediction}. We employ a simple deconvolution layer on the object-conditioned image feature $\hat{F}_i^{k}$ to obtain the heatmap of keypoint and mask of the object $i$ in each perspective $k$.


\textbf{Rotation prediction} The rotation angle is represented as a 2D vector representing the sine and cosine values of it. The rotation predictor accepts conditional image features $\hat{F}_i^{k}$ as input and outputs the sine and cosine value.

\textbf{Confidence score prediction} Since bricks may not be visible at all perspectives, here we predict the the confidence score $c_i^{k}$ of each brick at each perspective. Specifically, $c_i^{k}$ represents the Iou(Intersection Over Union) between the predicted mask and ground truth mask.

\begin{wrapfigure}{r}{0.48\textwidth}
  \begin{center}
  \includegraphics[width=0.75\linewidth]{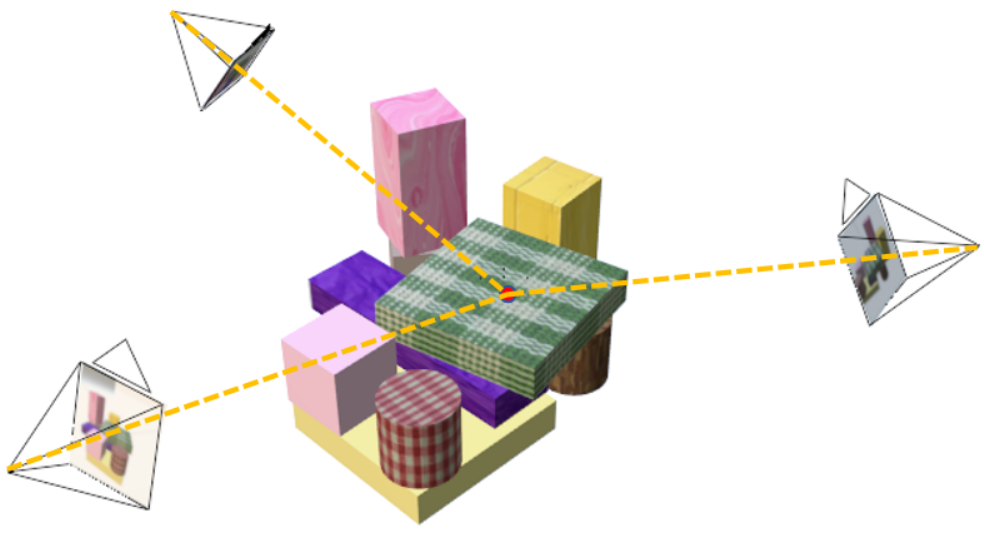}
  \end{center}
  \caption{\small Illustration of 3D position prediction module.}
  \label{fig:pose_estimation}
\end{wrapfigure}

During inference, the pose of each object in 3D space is obtained by merging the poses from each perspective (see Figure \ref{fig:pose_estimation}). In more details, for 3D position prediction, our method involves detecting keypoints of the object parts from the view whose confidence score $c_i$ is higher than a threshold $\theta$. Then, utilizing the camera parameters, the rays in 3D space generated by keypoints are used to infer the object's position in 3D space. Each ray $R_i$ is represented as $r_i(t) = O_i + t \cdot D_i$ where $O_i$ is the origin and $D_i$ is the direction.
Our objective is to find a point $P$ that minimizes the function:
\begin{eqnarray}
h(Z) = \sum_i^{L} d(Z, R_i),
\end{eqnarray}
where $L$ is the number of rays and $d(Z, R_i)$ represents the shortest distance from the point $P$ to the ray $R_i$.  Here, the minimization of the objective function $h(Z)$ is approached through the gradient descent method.



\subsection{Training and Loss functions}
We train Neural Assembler with full supervision on the generated dataset where each sample we have the groundtruth shape, texture, keypoint, mask, rotation information of each brick, the number of bricks and the relationship graph of bricks. The entire neural network is trained end-to-end with gradient
descent. Our objective function is computed by $L=\alpha \cdot L_{count} + \beta \cdot L_{graph} + L_{pose}$, where $L_{count}$ is the L1 Loss between the predicted number of bricks and ground truth $count_{gt}$.

Following \cite{carion2020end}, bipartite matching is used to find an optimal permutation $\{\sigma_i\}_{i=1}^N$ to match the N object queries and ground truth bricks in the scene. The pose loss of bricks includes the loss of shape, texture, keypoint, mask and rotation.
\begin{align}
L_{pose} &= L_{keypoint} + L_{mask} + {\gamma_1}L_{rotation}\\
&+ {\gamma_2}L_{shape} + {\gamma_3}L_{texture} + {\gamma_4}L_{confidence},
\end{align}
where $L_{keypoint}$ is the focal loss \cite{lin2017focal} computed based on the predicted heatmap and ground truth heatmap generated by $Kps_{\sigma_i}$, $L_{mask}$ is the focal and dice loss between the predicted mask and ground truth mask $M_{\sigma_i}$, $L_{rotation}$ is the L1 Loss between the prediced sine and cosine and the ground truth sine and cosine of $Rot_{\sigma_{i}}$, $L_{shape}$ and $L_{texture}$ are the cross entropy loss for shape and texture classification and $L_{confidence}$ is L1 Loss between the predicted confidence score and Iou of the predicted mask and ground truth mask. 
Our model strategically prioritizes the hyperparameters $L_{keypoint}$ and $L_{mask}$ due to their critical impact on object detection, essential for accurate object interaction and identification in complex scenes. In contrast, $L_{rotation}$, $L_{shape}$, $L_{texture}$ and $L_{confidence}$ are assigned a reduced weight of 0.1 each, a decision grounded in empirical findings that highlight their relatively minor incremental benefits to overall model efficacy.

$L_{graph}$ is the loss for relationship graph prediction. Firstly, we define the loss for any subset of the entire edge set. For a subset $\hat{E}$ of the edge set $E$ of the complete graph, the edge loss of $\hat{E}$ is defined as $L_{\hat{E}}=\sum_{(x,y)\in \hat{E}} L_{CE}(P_{(x,y)}, \hat{E}_{(x,y)})$. Then the $L_{graph}$ is defined as $L_{graph}=L_{E}+L_{topK_E}$ 
\begin{eqnarray}
L_{topK_E}=\frac{1}{K_E}\sum_{k=1}^{K_E} L_{E_{top_k}},
\end{eqnarray}
where $E_{top_k}$ is the set of the edges with the top $k$ highest predicted probability. Since the entire relationship graph is a directed graph with sparse edges, the hyperparameter $K_E$ is defined as $count_{gt} + 1$.





\section{Experiments}

\textbf{Dataset preparation}. Experiments are conducted on two self-constructed datasets. The CLEVR-Assembly Dataset, created via the CLEVR-Engine\cite{johnson2017clevr}, comprises a shape library with 6 brick shapes and 16 textures, a 76.5\% visibility probability of each brick per perspective, 7.51 bricks per sample, and an average assembly graph depth of 4.01, with approximately 10K training, 500 validation, and 2000 test samples. The LEGO dataset, synthesized using Pytorch3d, features 12 LEGO brick shapes and 8 textures, an 82.6\% visibility probability of each brick per perspective, 7.39 bricks per sample, and an average graph depth of 4.49, also with approximately 10K training, 500 validation, and 2000 test samples. The two datasets, characterized by brick number, occlusion from variable visibility, and complex assembly graph, reflect the complexity of assembly tasks. 

\textbf{Baseline models}. In addressing this novel task, for which no direct baseline exists, we have established a comparative framework against three distinct baseline methods to demonstrate the efficacy of our approach. First, to assess the validity of our assembly order prediction methodology, we introduce a Long Short-Term Memory (LSTM) \cite{graves2012long} module as a surrogate baseline to contrast with our Graph Convolutional Network (GCN) based module. This comparison aims to highlight the enhanced predictive capabilities our GCN model brings to complex assembly sequences.

Furthermore, for the object pose estimation component, our methodology is rigorously benchmarked against DETR3D~\cite{wang2022detr3d}, a prominent baseline in the realm of object detection within autonomous driving scenarios. This comparison is pivotal in underscoring the adaptability and accuracy of our model in 3D pose estimation, a crucial aspect in varied application domains.

Lastly, in evaluating our multi-view image feature fusion process, we contrast our approach with a method that does not leverage scene consensus. This comparison is instrumental in showcasing the enhanced scene understanding and feature integration our method offers, thus demonstrating its superiority in synthesizing and utilizing multi-view image data.

\textbf{Implementation details}. Our approach is implemented in single-scale version for fair comparison with other works. It incorporates a CLIP\cite{radford2021learning} pre-trained ViT-B/16 image encoder, a PointNet-based~\cite{Qi_2017_CVPR} point cloud encoder, and a ResNet-18\cite{he2016deep} for texture encoding. We employ a two-layer residual network for brick number prediction. The shape, material, iou prediction heads are implemented using 3-layer MLP and ReLU activations. Rotation prediction also uses a two-layer residual network, and our GCN architecture employs two message-passing layers. Training is conducted on an RTX 3090 GPU using AdamW, with an initial rate of 5e-4, decaying by 0.8 per epoch, a weight decay of 1e-3, and batch size 8 over 10 epochs for both datasets.

\textbf{Evaluation metrics}. In this assembly task, we introduce several metrics to evaluate the performance of our algorithm both at a per-scene and per-step level, providing a holistic measure of our method's efficacy. Specifically, for the per-scene metrics, our approach necessitates the prediction of the entire assembly sequence based on multi-view images, emphasizing the ability to comprehend and reconstruct the complete scene from various perspectives. In contrast, the per-step metrics operate under the assumption that the assembly order is known a priority. Here, we focus on calculating the error between the predicted information for each individual brick and the corresponding ground truth, independent of the assembly order. This allows for a comprehensive evaluation of the method's ability in both holistic scene understanding and step-wise brick analysis.

\begin{table}[h]
  \centering
  \caption{Comparison of per-scene metrics. } 
  \begin{scriptsize}
  \begin{tabular}{cccccc}
    \toprule
     \multirow{1}{*}{Method} & Complete Rate & Per-scene Acc & Count Acc & Order CR \\
    \midrule 
    \multicolumn{5}{c}{LEGO-Assembly} \\
    \midrule
    LSTM~\cite{graves2012long} & 27.5 & 5.3 & 60.3 & 35.1 \\
         DETR3D~\cite{wang2022detr3d} & 25.8 & 2.5 & 61.5 & 63.5 \\
        Ours (w/o consensus)  & 43.7 & 18.4 & 69.0 & 64.5 \\
        Ours   & 43.9 & 22.9 & 76.3 & 69.4 \\
    \midrule 
    \multicolumn{5}{c}{CLEVR-Assembly}\\
    \midrule
     LSTM~\cite{graves2012long} & 19.7 & 8.0 & 91.5 & 22.6 \\
          DETR3D~\cite{wang2022detr3d} & 16.8 & 4.5 & 89.5  & 35.3 \\
        Ours (w/o consensus)  & 28.6 & 6.6 & 92.1 & 45.5 \\
        Ours (2 views) & 22.0 & 4.6 & 88.7 & 38.6 \\
         Ours (3 views) & 25.7 & 9.3 & 94.0  & 44.5 \\
        Ours   & 41.5 & 22.5 & 95.5 & 62.1 \\
    \bottomrule
  \end{tabular}
  \end{scriptsize}
  \label{Per-scene}
  
\end{table}

For per-scene metrics, we evaluate the Complete Rate (completion percentage of the brick model), Order CR (completion rate of the sequence of brick types), Per-scene Acc (accuracy of completely assembling an entire brick model), Count Acc(precision of predicting the number of bricks). For per-step metrics, we evaluate the Pos Acc and Rot Acc (3D position accuracy and rotation accuracy), Shape Acc and Texture Acc (shape accuracy and texture accuracy), Kps Mse (error of the predicted 2D keypoints of the object), mIou (mean Intersection over Union between the predicted mask and the ground truth), the F1-score between predicted relation graph and ground truth relation graph and Per-step Acc(accuracy of correct predictions for each brick's information).



\textbf{Results on CLEVR-Assembly}. Per-scene quantitative results on the CLEVR-Assembly Dataset are summarized in Table \ref{Per-scene}. Neural Assembler outperforms baseline models in all metrics considered. From Table \ref{Per-Step}, we can see Neural Assembler locates objects more accurately than DETR3D. 

\begin{wrapfigure}{r}{0.5\textwidth}
\begin{center}
\resizebox{\linewidth}{!}{
\includegraphics[width=\linewidth]{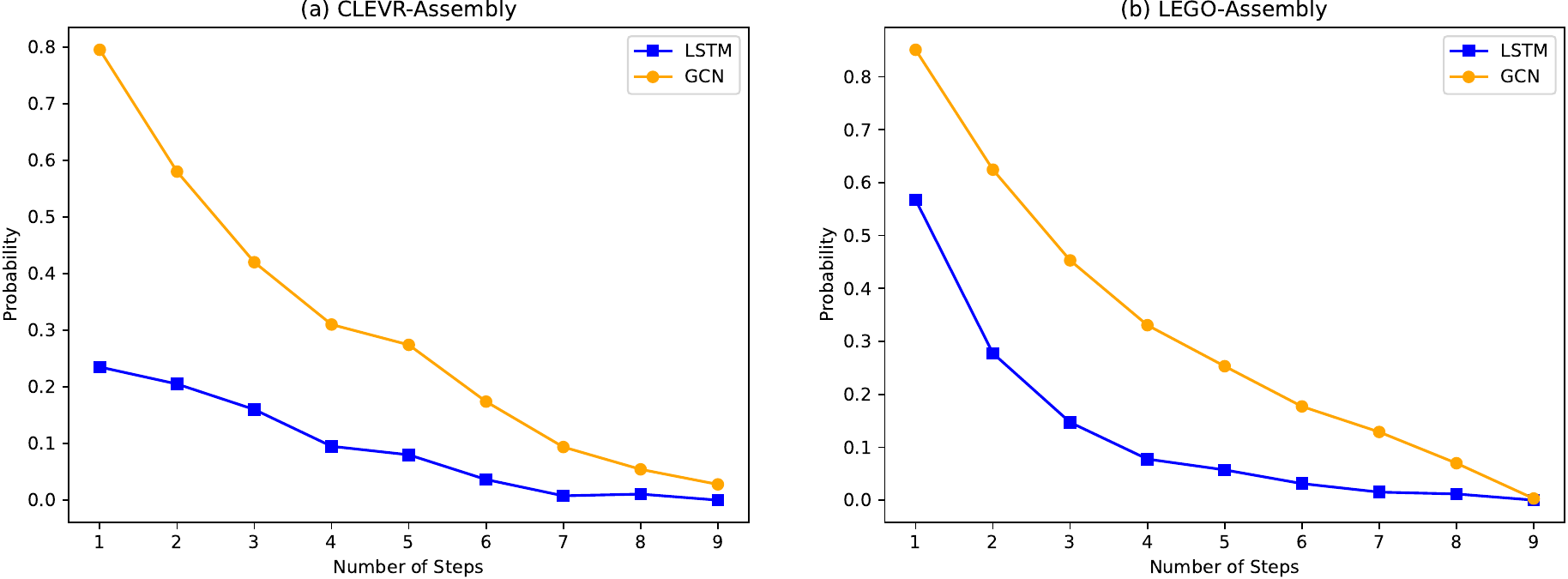}
}
\end{center}
   \caption{\small The probability distribution of CCA.}
\label{fig:lstm}
\end{wrapfigure}

The metric CCA proposed by \cite{chen2019order} is adopted here for the brick order evaluation. It denotes the probability distribution of the number of brick that a model can consecutively place from scratch. As shown in Fig.\ref{fig:lstm}, LSTM perform worse than GCN. This is because time dependence is not crucial for the assembly order prediction. Instead, the assembly problem requires prediction from complex spatial relationships. The adeptness of GCN in capturing spatial relation plays a critical role in understanding the assembly order.





\textbf{Consensus Module} As shown in Table \ref{Per-Step}, extracting the consensus can better align the information of the images from various perspectives. Without scene consensus, it is difficult for the model to integrate information from multi-view images to obtain the overall information of each brick.

\textbf{Number of views} Furthermore, we compared the results obtained by accepting  different numbers of images as input. As shown in Tables~\ref{Per-scene} and \ref{Per-Step}, the result shows that more perspectives as input can improve the performance. This is because each brick may not be seen from some perspectives due to the existence of occlusion. More views mean more information for prediction.

\begin{figure}[t]
  \centering
  \begin{minipage}{0.48\textwidth}
    \includegraphics[width=\linewidth]{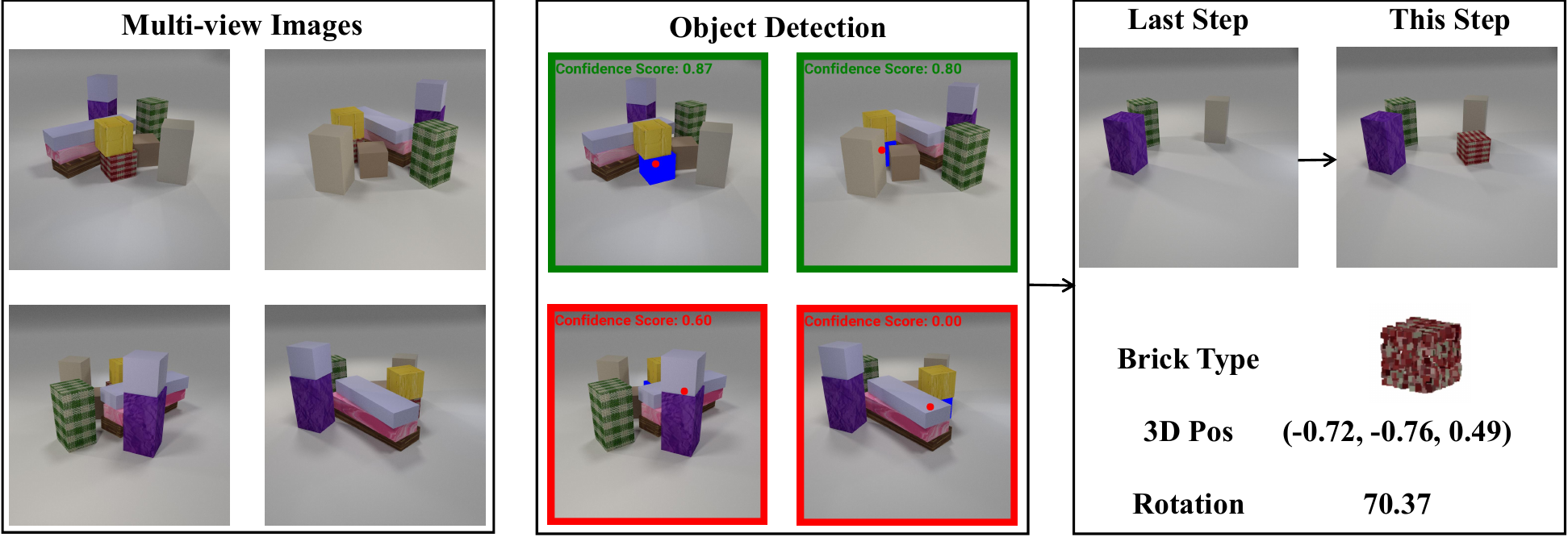}
    \caption{\small Result from CLEVR-Assembly Dataset.}
    \label{fig:clevr_assembly}
  \end{minipage}
  \hfill 
  \begin{minipage}{0.48\textwidth}
    \includegraphics[width=\linewidth]{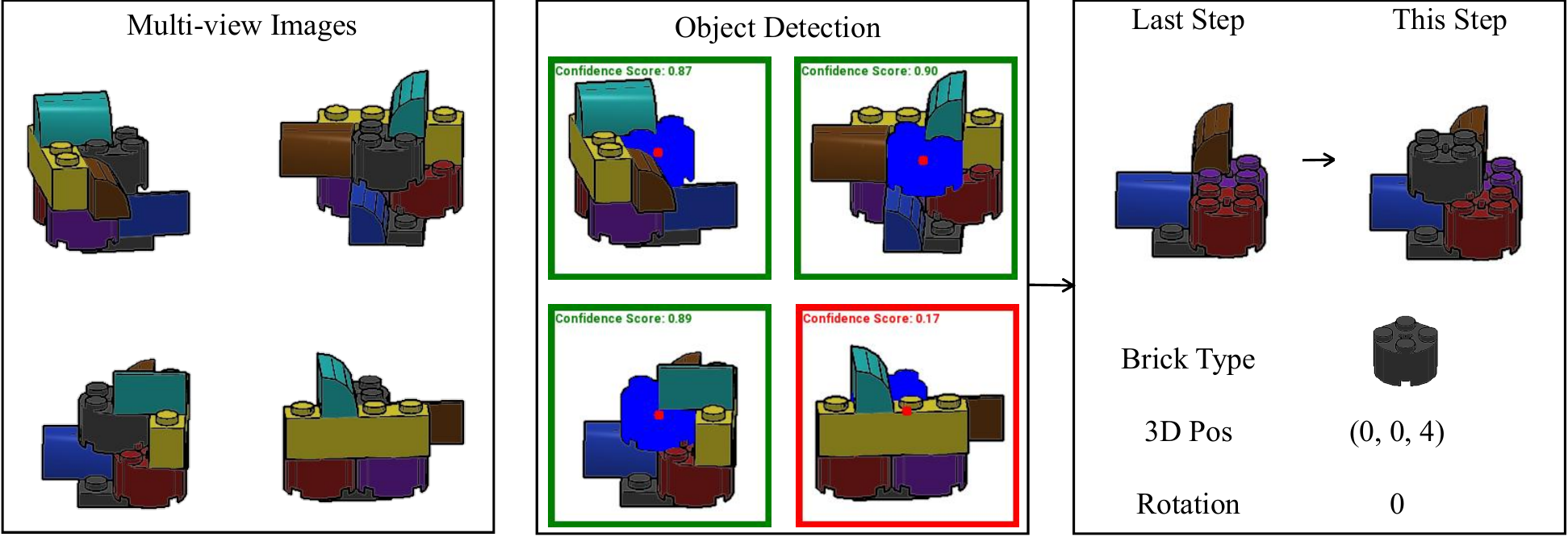}
    \caption{\small Result from LEGO-Assembly Dataset.}
    \label{fig:LEGO_assembly}
  \end{minipage}
\end{figure}

\begin{table}[h]
  \centering
  \caption{Comparison of baselines on per-step metrics.} 
  \begin{scriptsize}
  \begin{tabularx} {\linewidth}{ccccccccc}
    \toprule
    \multirow{1}{*}{Method} & Per-Step Acc & Pos Acc & Rot Acc & Shape Acc & Texture Acc & mIoU & Kps Mse & F1  \\
    \midrule
    \multicolumn{9}{c}{LEGO-Assembly}\\
    \midrule    
     DETR3D~\cite{wang2022detr3d} & 41.7 & 47.2 & 78.2 & 87.8 & 98.3 & - & - & 0.797   \\
       Ours (w/o consensus)  & 71.7 & 79.0 & 87.2 & 89.9 & 98.3 & 76.3 & 1.12 & 0.829   \\
       Ours   & 73.5 & 80.4 & 88.0 & 91.5 & 98.3 & 78.5 & 0.88 & 0.820  \\
    \midrule
    \multicolumn{9}{c}{CLEVR-Assembly}\\
    \midrule
    DETR3D~\cite{wang2022detr3d} & 29.2 & 32.4 & 75.6 & 72.4 & 67.0 & - & - & 0.734 \\
       Ours (w/o consensus)  & 57.2 & 67.9 & 78.9  & 87.1  & 88.5 & 61.2 & 1.2 & 0.781 \\
       Ours (2 views) & 56.1 & 71.7 & 73.0 & 80.0 & 86.4 & 65.4 & 2.5 & 0.721 \\
       Ours (3 views) & 61.1 & 74.6 & 77.9  & 85.7  & 90.7 & 67.1 & 1.5 & 0.772 \\
       Ours  & 69.2 & 79.1 & 84.1 & 91.5  & 93.8 & 71.1 & 0.78 & 0.840 \\
    \bottomrule
  \end{tabularx}
  \end{scriptsize}
  \label{Per-Step}
\end{table}

Fig.~\ref{fig:clevr_assembly} further shows the generated assembly instructions for a brick model in the CLEVR-Assembly Dataset. Perspectives with confidence score greater than 0.66 are selected to infer the brick's information. It is evident that Neural Assembler is adept at excluding perspectives where bricks are obscured by predicting confidence scores, thereby identifying optimal perspectives for predicting positional information. Concurrently, it is capable of predicting the structure between bricks to determine the appropriate assembly sequence.

\textbf{Results on LEGO-Assembly}. Different from CLEVR-Dataset, LEGO bricks are connected through slots. It is easier to infer the position of the bricks based on the connection constraints between the LEGO bricks, as long as the rough position of the LEGO bricks is predicted. However, there are many challenges in predicting the assembly sequence of LEGO brick models. For instance, the more compact assembly of LEGO bricks results in increased occlusion.

The LEGO brick will only have rotations chosen from
$(0^{\circ}, 90^{\circ} , 180^{\circ}, 270^{\circ})$. Meanwhile, the position of LEGO bricks is discretized. We adopt the connection-constrained inference subroutine and an inference-by-synthesis subroutine used in \cite{wang2022translating} to predict the position and rotation angle for each view, and then integrated them through voting. The results in Table \ref{Per-scene} and Table \ref{Per-Step} shows that Neural Assembler can yield more
accurate results than other baselines.
Fig.\ref{fig:LEGO_assembly} further shows the generated assembly instructions for a LEGO model.

\begin{table*}[h]
  \centering
  \begin{tabular}{ccc}
    \toprule
    Method & Per-scene Acc & Complete Rate \\
    \midrule
    \multicolumn{3}{c}{Novel-Dataset}\\
    \midrule
    LSTM~\cite{graves2012long} & 16.0 & 27.3 \\
    DETR3D~\cite{wang2022detr3d} & 7.3 & 21.8 \\
    Ours & 34.2 &  58.5 \\
    \midrule
    \multicolumn{3}{c}{Real-World Dataset}\\
    \midrule
    LSTM~\cite{graves2012long} & 7.3 & 21.8 \\
    DETR3D~\cite{wang2022detr3d} & 2.4 & 12.8 \\
    Ours & 22.0 & 50.5 \\
    \bottomrule
  \end{tabular}
  \caption{The performance of the fine-tuned model on the novel simulated dataset and real-world dataset.}
  
  \label{ft}
  
\end{table*}

\begin{figure}[h]
\begin{center}
\includegraphics[width=\linewidth]{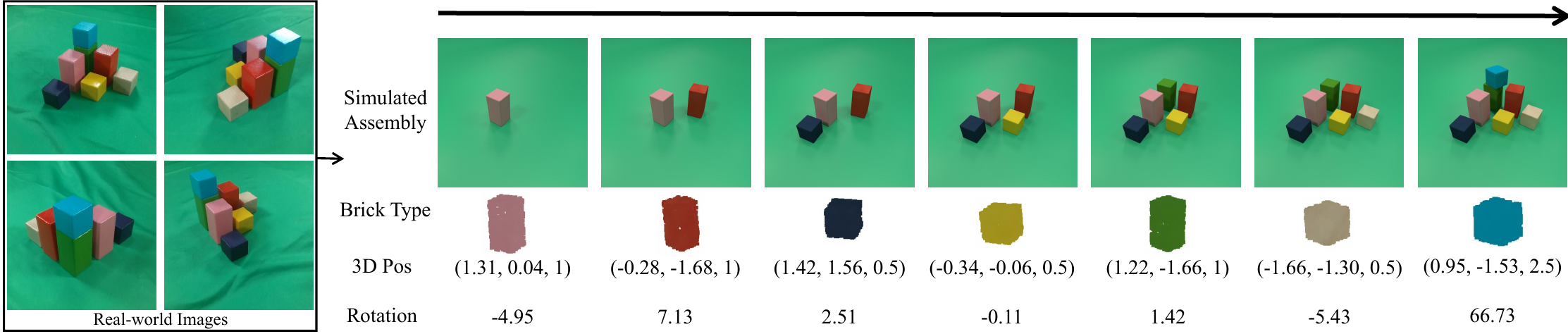}
\end{center}
   \caption{\small The result from the real-world brick model. The left box displays 4 images captured using a Realsense camera, while the right delineates the detected type, position, rotation angle of each brick, and the sequential assembly order of the brick model.}
\label{fig:demo_assembly}
\vspace{-0.05in}
\end{figure}
\textbf{Real-world experiments}. 

To confirm the model's generalizability, a comprehensive test dataset is constructed, complete with annotations for each brick’s shape, position, and rotation. For each sample, we acquired real brick images using a Realsense camera in the real world and generated corresponding simulated images in the simulation environment employing real-world camera parameters to ensure the consistent coordinate between the simulated and real environments. The dataset encompasses 5 brick types and 7 textures, averaging 6.1 bricks per brick model.


To evaluate the Neural Assembler, we collected point clouds and textures from real bricks. This data facilitated the creation of a synthetic dataset, used for fine-tuning the model initially trained on the CLEVR-Assembly dataset.

As indicated in Table~\ref{ft}, the Neural Assembler achieves performance in real-world experiments close to the results obtained in simulated environments, demonstrating its robust applicability. Fig.~\ref{fig:demo_assembly} presents the result on the real world dataset.

\begin{figure}[h]
\begin{center}
\vspace{-10pt}
\includegraphics[width=0.95\linewidth]{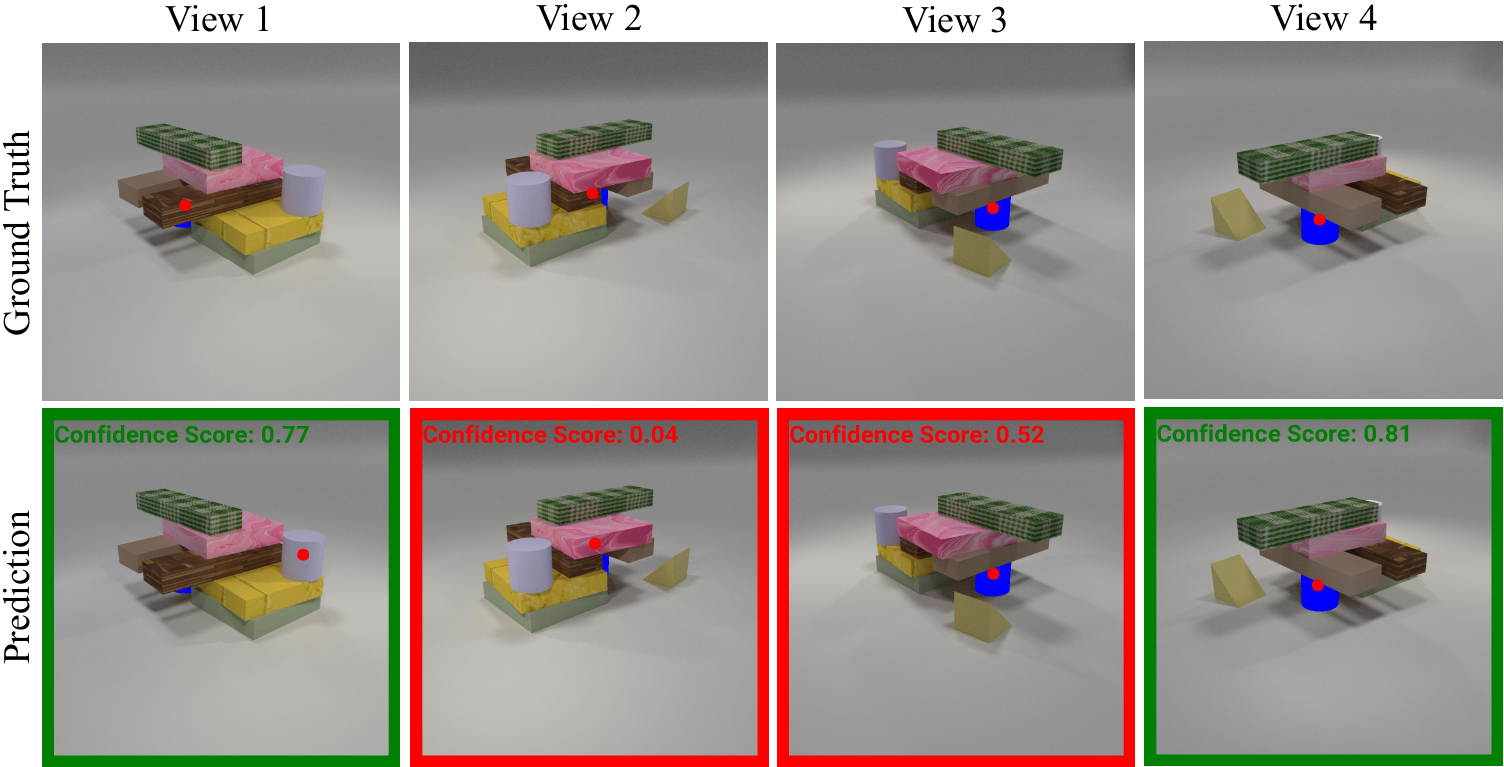}
\caption{\small Failure case. The model confidently but incorrectly predicts the highlighted block in View 1, while in View 3, despite correct keypoint identification, occlusion results in a less confident. This causes erroneous overall prediction.}
\label{fig:failure_1}
\end{center}
\vspace{-0.1in}
\end{figure}

\textbf{Discussion}. As shown in Fig.~\ref{fig:failure_1}, the occlusion still greatly affects the performance of the model, especially the objects under the brick model will be greatly blocked by the bricks pressing on it. To alleviate this problem, in future work, we plan to enhance model performance with a deeper integration of physical scene understanding. The model is expected to not only interpret the visual aspects but also the underlying physical principles governing the scene. 

\section{Conclusion}
We study the problem of generating robotic assembly instructions from multi-view images and propose Neural Assembler, a model that predicts the assembly instructions of the brick model. The key idea behind our model is to learn the graph structure to predict the relationships among bricks and infer the 3D pose according to multi-view images. Results show that our model outperforms existing methods on the newly collected CLEVR-Assembly and LEGO-Assembly Dataset.

\medskip

{
\small

\bibliographystyle{ieeenat_fullname}
\bibliography{neurips_2024}


}


\appendix

\section{Appendix / supplemental material}

The data generation pipeline for CLEVR-Assembly and LEGO-Assembly dataset is introduced in Section \ref{sec:dataset}. Then we provide more implementation details and hyperparameters of models in Section \ref{sec:model}. Section \ref{sec:real} shows the details of the real-world robotic experiment. 

\subsection{Dataset Generation}
\label{sec:dataset}
\textbf{CLEVR-Assembly Dataset} In constructing the CLEVR-Assembly dataset, each sample is randomly generated. For each step of assembly, a brick with a random shape and texture is selected, followed by randomizing its horizontal position $(x, y)$ and rotation angle. The operation is rolled back if the brick is unstable upon free fall. Each brick's information including keypoints, mask, 3D coordinate, rotation, shape, and texture is recorded, and relationships with other bricks are computed. All models are confined within a $[-3, 3]\times[-3, 3]$ horizontal area, with the smallest cubic brick being $1\times1\times1$ in size. 

The CLEVR-Assembly dataset comprises various brick shapes: cubes $(1\times1\times1)$, rectangular prisms $(1\times1\times2,\ 1\times2\times0.5,\ 2\times2\times0.5)$, cylinder with a base diameter of 1 unit and a height of 1 unit, and triangular prisms featuring a square base with each side measuring $\sqrt{2}$ units and a height of $\frac{\sqrt{2}}{2}$ units. For textures, 8 distinct brick textures are sourced, and their average colors are used, leading to 16 unique textures. This results in a total of 96 different brick categories.

\textbf{LEGO-Assembly Dataset} 
$\textbf{LEGO-Assembly Dataset}$ For the LEGO-Assembly dataset construction, each sample begins with selecting a base brick randomly. Subsequently, at every step, we choose a random brick and determine feasible poses for this brick, considering the connection constraints of the pre-assembled brick model. A pose is then randomly selected from this set as the brick's final placement.

The LEGO-Assembly dataset selects 12 distinct brick types from the LEGO library and utilizes 8 different colors for textures, resulting in a total of 96 unique brick categories.

Four cameras are positioned at the left-front, left-back, right-front, and right-back of the brick model. Their translations are randomly chosen within a sphere of radius 1.5 at a distance of 12 from the origin. The rotation is sampled from $(0, (90\cdot k \pm 30 )^{\circ}, (45 \pm 15)^{\circ})$ for camera $k$.

\subsection{Implementation Details}
\label{sec:model}
\textbf{Hyperparameters} For training loss:
\begin{align}
L &= \alpha \cdot L_{count} + \beta \cdot L_{graph} + L_{pose},\\
L_{pose} &= L_{keypoint} + L_{mask} + {\gamma_1}L_{rotation}\\
&+ {\gamma_2}L_{shape} + {\gamma_3}L_{texture} + {\gamma_4}L_{confidence},
\end{align}
We set $\alpha$ to 0.2, $\beta$ to 0.5, $\gamma_1$ to 0.1, $\gamma_2$ to 0.1, $\gamma_3$ to 0.1 and $\gamma_4$ to 0.1.
All models are trained using AdamW, with an initial rate of 5e-4, decaying by 0.8 per epoch, a weight decay of 1e-3, and batch size 8 over 10 epochs for both datasets. We use the pre-trained ViT-B/16 weights and fine-tune it with the learning rate setting to the same value as other modules.

\textbf{Details of 3D Pose Inference}  The center of each brick is defined as the keypoint. For 3D pose estimation, we select the perspective with a confidence score greater than a threshold and extract its 2D keypoint coordinates 
$(x,y)$. In the camera coordinate system, we derive the world coordinates for points $(x,y,-1)$ and $(x,y,1)$ in the camera coordinate, which form the predicted ray for the keypoint from the perspective. Consequently, the 3D position of the keypoint is determined by the intersection of rays predicted from multiple perspectives.

\textbf{Model Architecture} 
A pre-trained Vision Transformer (ViT-B/16) processes an image of size $224\times 224$, yielding image features of dimension $768\times(196+1)$. These features are then transformed via a fully connected layer into a feature space of $256\times(196+1)$, where 196 represents the number of tokens, equating to $14\times14$, and the resulting CLIP feature vector has a dimension of $1\times256$. Concurrently, the PointNet processes the point cloud of size $N_1\times1024\times3$ to extract $N_1\times256$ shape features of the brick. For texture information, $N_2$ images of size $224\times224$, processed through the ResNet-18 network, yield the texture features of dimension $N_2\times256$.

For the Transformer Decoder, the query is constituted by object queries of dimension $(N_1+N_2+16)\times256$. The key is established as $196\times256$ image embeddings, augmented with positional encoding, while the value is set as the image embeddings, obtaining the global object features. The per-view object feature and the fused image feature  are intricately processed by  the Linear Interpolation module to yield the object-conditioned image feature $\hat{F_i^k}$ of dimension $256\times14\times14$.

\subsection{The computational complexity}
The runtime of the baseline consists of forward propagation (T1)
and inferring bricks’ poses and assembly sequence (T2). The
average T1 for Neural Assembler/ DETR3D / LSTM is 0.24s
/ 0.23s / 0.20s respectively. For T2, the averages are 1.79s /
1.04s / 1.32s for Neural Assembler / DETR3D / LSTM respectively. The total runtime T1 + T2 is acceptable.

\begin{figure}[h]
\begin{center}
\includegraphics[width=\linewidth]{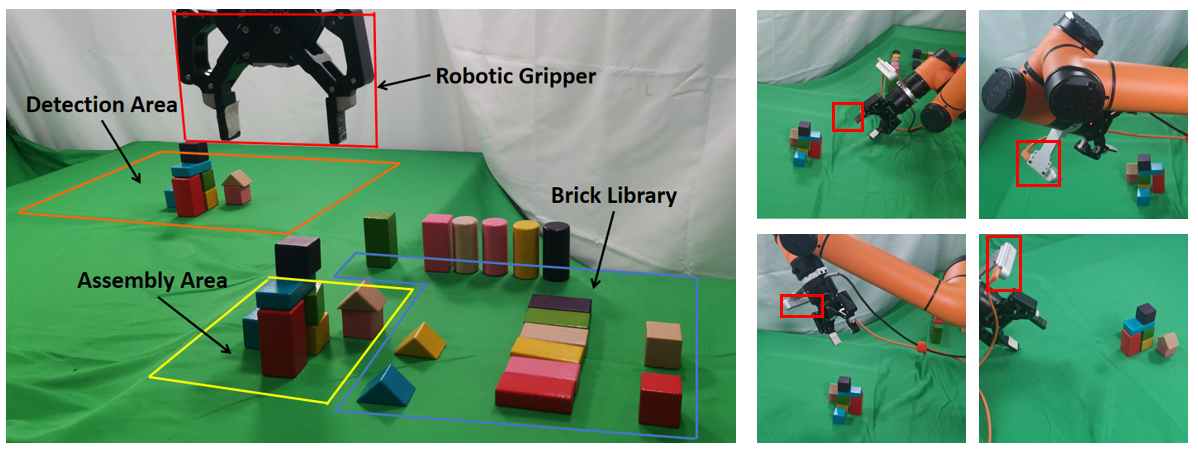}
\end{center}
\caption{Real-world robotic experiment scenario. The image above shows the assembly scene, including Detection Area, Assembly Area, Robotic Gripper, Brick Library. In the bottom four images, the red box within each view represents  camera on the Robotic Gripper.}
\label{fig:scene}
\vspace{-0.05in}
\end{figure}

\subsection{Real-World Robotic Experiment}

\label{sec:real}
To evaluate the practical applicability of Neural Assembler, a real-world assembly experiment is conducted. The experimental setup comprises an Aubo-i5 robotic arm equipped with an Intel RealSense D435i RGB-D camera, facilitating precise visual perception. The manipulation component involves a Robotiq 2F-85 two-finger gripper, providing adept grasping capabilities.

The grasping process begins as the robotic arm captures images from four distinct perspectives within the Detection Area, as illustrated in Fig. \ref{fig:scene}. This enables the extraction of structural information about the brick model. Subsequently, the arm relocates to the Assembly Area, where it utilizes bricks from the Brick Library to reconstruct the brick model. At each step of assembly, the arm determines the brick's location within the library based on its shape and texture. The brick is then grasped from a vertical direction and positioned at the predicted pose. 

\end{document}